\documentclass[conference]{IEEEtran}
\IEEEoverridecommandlockouts
\usepackage[T1]{fontenc}
\usepackage[greek, english]{babel}
\usepackage{color}%
\usepackage[pdftex]{graphicx}
\usepackage{random}
\usepackage{amsfonts} 
\usepackage{amssymb} 
\usepackage{amsmath} 
\usepackage{amsbsy} 
\usepackage{fixmath}
\usepackage{import}
\usepackage{float}

\ifCLASSINFOpdf
\else
\fi
\begin{document}
%
\title{Predicted-occupancy grids for vehicle safety applications based on autoencoders and the Random Forest algorithm%
\thanks{©~2017 IEEE. Personal use of this material is permitted. Permission from IEEE must be obtained for all other uses, in any current or future media, including reprinting/republishing this material for advertising or promotional purposes, creating new collective works, for resale or redistribution to servers or lists, or reuse of any copyrighted component of this work in other works.
This is the author's accepted manuscript version of the paper published in the 2017 International Joint Conference on Neural Networks (IJCNN), Anchorage, AK, USA, pp.~1244--1251, 2017. DOI: 10.1109/IJCNN.2017.7965995.}}

\author{\IEEEauthorblockN{Parthasarathy Nadarajan}
\IEEEauthorblockA{ Technische Hochschule Ingolstadt \\
Ingolstadt, Germany\\
Email: parthasarathy.nadarajan@thi.de}
\and
\IEEEauthorblockN{Michael Botsch}
\IEEEauthorblockA{Technische Hochschule Ingolstadt \\
Ingolstadt, Germany\\
Email: michael.botsch@thi.de}
\and
\IEEEauthorblockN{Sebastian Sardina}
\IEEEauthorblockA{RMIT University\\
Melbourne, Australia\\
Email: sebastian.sardina@rmit.edu.au
}}

%


\maketitle

\begin{abstract}
In this paper, a probabilistic space-time representation of complex traffic
scenarios is predicted using machine learning algorithms. Such a representation is significant for all active vehicle safety applications especially when performing dynamic maneuvers in a complex traffic scenario. 
As a first step, a hierarchical situation classifier is used to
distinguish the different types of traffic scenarios. This
classifier is responsible for identifying the type of the road
infrastructure and the safety-relevant traffic participants of the
driving environment. With each class representing similar traffic
scenarios, a set of \emph{Random Forests} (RFs) is individually
trained to predict the probabilistic space-time representation,
which depicts the future behavior of traffic participants. This
representation is termed as a \emph{Predicted-Occupancy Grid}
(POG). The input to the RFs is an \emph{Augmented Occupancy Grid}
(AOG). In order to increase the learning accuracy of the RFs and
to perform better predictions, the AOG is reduced to
low-dimensional features using a \emph{Stacked Denoising
Autoencoder} (SDA). The excellent performance of the proposed
machine learning approach consisting of SDAs and RFs is
demonstrated in simulations and in experiments with real vehicles.
An application of POGs to estimate the criticality of
traffic scenarios and to determine safe trajectories is also
presented.
\end{abstract}


%
\IEEEpeerreviewmaketitle

\section{Introduction}

A new generation of active safety systems has appeared on the
market due to an improved environment detection technology and
situation assessment capabilities~\cite{Winner16}. These systems
are responsible for the avoidance and mitigation of collisions,
e.\,g., \emph{Autonomous Emergency Braking}
(AEB)~\cite{Kaempchen09}. One of the challenges faced by
these systems is to understand an encountered traffic scenario by
considering vital information such as the road infrastructure,
relevant traffic participants and their corresponding future
behaviors.

Several research is currently being done to anticipate the behavior of the traffic participants for specific situations such as at
intersections~\cite{Lefevre11} and at traffic
lights~\cite{Aoude11}. In~\cite{Nadarajan16}, a novel approach was
presented for the estimation of how a particular complex traffic
scenario with multiple objects will evolve in the future using the \emph{Random Forest} (RF) algorithm~\cite{Breiman01}. An efficient
space time representation of the future traffic scenario, namely \emph{Predicted Occupancy Grid} (POG) was
formulated. However, the above mentioned approaches are applicable only
for a specific configuration of a traffic scenario and hence are not
general enough. In order to address this issue, it is important to
have a refined methodology capable of extracting the relevant
information about the driving environment and its participants, and to
learn and use that information for predicting the behavior of the traffic
participants when a similar scenario is
encountered~\cite{Armand13}. In~\cite{Bonnin12}, a generic
solution to predict the behavior of the surrounding vehicles on a
large variety of scenes based on classification was presented.
Also, a machine learning method based on semantic reasoning was
proposed in~\cite{Dianov15} to detect and extract meaningful
information from different traffic scenarios and to infer the
correct driving behavior of the traffic participants. Similarly,
ontology based approaches were also used for analyzing traffic
scenarios~\cite{Hulsen11}.

In this paper, a ``Divide and Conquer'' approach is proposed to
identify different kinds of traffic scenarios, its meaningful
traffic participants and to predict the future driving behaviors
of the selected traffic participants. In order to handle a large
number of different traffic scenarios, a hierarchical classifier
with two levels is constructed. The first level is responsible for
identifying the type of the road infrastructure, e.\,g., straight
road, curved road, junction, etc. and the second level identifies
the safety-relevant traffic participants in the traffic scenario
under consideration. Each leaf of the hierarchical classifier will
correspond to a particular traffic scenario and an RF algorithm
will be specifically trained for that particular leaf to predict
the behavior of the traffic participants.

The input to the machine learning algorithm employed
in~\cite{Nadarajan16} was termed as the \emph{Augmented Occupancy Grid} (AOG).
The cells in such occupancy grid are augmented with
information about the road infrastructure and the traffic
participants such as acceleration, velocity and yaw angle. Because
of this augmentation, the AOG turns out to be
a high dimensional vector. It has been proven that performing machine learning on such high dimensional data is difficult~\cite{Meng16}.
Hence, an efficient representation of the data is important for
all machine learning and big data approaches. Using a deep
learning approach to extract the features is a well known
procedure. In~\cite{Hinton06}, high dimensional vectors were
converted to low dimensional vectors by training a multi-layered
neural network. It has also been shown in~\cite{Vincent08}
and~\cite{Taylor07} that autoencoders and Restricted Boltzmann
machines are capable of retrieving relevant features in an
unsupervised manner, respectively. In~\cite{Liu14}, a deep sparse
autoencoder was employed to extract low dimensional features from
high dimensional human motion data and a random forest is used to
classify the low dimensional features representing human motion.
In this paper, the \emph{Stacked Denoising Autoencoder} (SDA)~\cite{Meng16} is used to extract robust low dimensional
features from the AOGs~\cite{Nadarajan16} in an unsupervised
manner and these features are in turn used by the RF algorithm to
predict the POGs.

The paper is organized as follows. Section \ref{secHSC} describes
the hierarchical methodology adopted for classifying different
kinds of traffic scenarios. Extraction of the low dimensional
features from the AOG using the autoencoders
and the estimation of POGs using the RFs is presented in Section
\ref{secPOG}. In Section \ref{secExp}, the evaluation of the
methodology with simulation results is presented. An application
of the proposed method in the field of vehicle safety is
demonstrated in Section \ref{secApplication}.

Throughout this work, vectors and matrices are denoted by lower
and upper case bold letters, respectively. A lower case bold letter represents a
column vector.

\section{Hierarchical situation classifier}
\label{secHSC} 
\begin{figure}[htb!]
    \begin{scriptsize}
    \vspace{-0.3cm}
    \centering
    \def\svgwidth{0.47\textwidth}
    \centerline{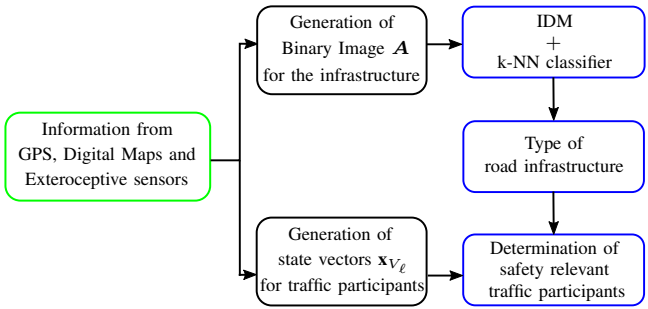}
    \end{scriptsize}
\def\figurename{Figure}
\caption{Hierarchical Situation Classifier.}
\label{figHSC}
\vspace{-0.2cm}
\end{figure}

This section explains the methodology for the
hierarchical classification of different traffic scenarios
encountered during driving. The main advantages of such a
hierarchy are that it facilitates a modular approach and handles newly encountered scenarios that do not match any of the predefined scenarios. Section
\ref{subsecIDRoad} shows the classification of different road
geometries with the help of an image matching algorithm and Section
\ref{subsecIDParticipants} explains the selection of relevant
traffic participants in a traffic scenario using a set of
predefined rules. An overview of the methodology can be seen in
the Figure~\ref{figHSC}.

The data from the sensors are transformed and represented in such
a way that the traffic scenario under observation is seen from the
perspective of the EGO vehicle, the vehicle in which the safety
system operates. This defines the coordinate frame for the
algorithm. The traffic area under consideration is $40$m $\times
40$m with the center of gravity of the EGO vehicle located
at $(2.5\text{m}, 0\text{m})$. Left side of the
Figure~\ref{figScenarioAssumption} depicts the
assumptions when the EGO vehicle (red) is driving towards a $3$
way intersection.

\subsection{Identification of the Road Infrastructure}
\label{subsecIDRoad} 
\begin{figure}[htb!]
    \begin{scriptsize}
    \vspace{-0.2cm}
    \centering
    \def\svgwidth{0.4\textwidth}
    \centerline{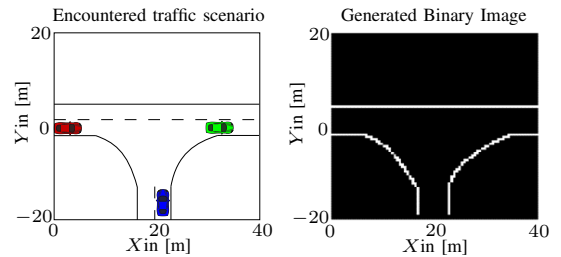}
    \end{scriptsize}
    \vspace{-0.2cm}
\def\figurename{Figure}
\caption{Generation of the binary image from a traffic scenario.}
\label{figScenarioAssumption}
\vspace{-0.2cm}
\end{figure}

The first level of the hierarchical situation
classifier is responsible for identifying the type of the road
geometry. The information about the road infrastructure is assumed
to be known from GPS, digital maps and exteroceptive
sensors. With this information, a binary $I \times J$ test image
$\boldsymbol{A} = \{a_{ij}\}$,  where $i = 1, \ldots, I $ and $j =
1, \dots, J $, of the road points is created. Such a binary image
is depicted in the right side of Figure
\ref{figScenarioAssumption} for the scenario shown in the left
side of the figure. In order to find the type of the road
infrastructure, an image matching approach is adopted. The image
$\boldsymbol{A}$ will be matched to one of the reference road
geometry templates $\boldsymbol{R}_k \in \{0,1\}^{M \times N}$,
with $k = 1, \ldots, K$ corresponding to different classes of road
geometries such as straight road, left curve, right curve, 3 way
intersection, etc. The $(m,n)$-th element of the matrix
$\boldsymbol{R}_k$ is $r_{mn}$. In this work, the dimensions of
the test image $\boldsymbol{A}$ and the reference image
$\boldsymbol{R}$ are the same, i.\,e., $I=M$ and $J=N$. For
performing the classification of the encountered road geometry to
one of the geometry templates $\boldsymbol{R}_k$, the \emph{Image
Distortion Model} (IDM) with the k-\emph{Nearest Neighbor} (k-NN)
classifier as described in~\cite{Keysers07} is employed.

\subsubsection{Image Distortion Model}
\label{subsubsecIDM} A detailed analysis on the different kinds of
image deformation models for image matching was performed
in~\cite{Keysers07}. Among the models, it was shown that the IDM
is capable of achieving high performance with low computational
complexity for real world image recognition tasks. The main aim of
the model is to find the optimal deformation from a set of
possible deformations in such a way that the distance between the
test and the reference image is the least. The IDM is a zero order
model where the relative displacement between the pixels are
disregarded and their absolute displacements are restricted.
Hence, mapping a test pixel $a_{ij}$ to a reference pixel $r_{mn}$
will not be more than $\Delta$ pixels from the place it would take
in a linear matching. With $m_{i} \in \{1, \ldots, M\} \cap
\{i-\Delta, \ldots, i+\Delta\}$ and $n_{j} \in \{1, \dots, N\}
\cap \{j-\Delta, \ldots, j+\Delta\}$, the IDM distance function
is defined as follows:
\begin{equation}
d(\boldsymbol{A},\boldsymbol{R}) = \sum_{i,j}{\min_{(m_{i},n_{j})}
d^{\prime}({a}_{ij},r_{m_{i}n_{j}})},
\end{equation}
where the local distance measure $d^{\prime}$ is the Euclidean
distance. The distance metric $d$ is used within the k-NN
classifier to obtain the class of the road infrastructure.

\subsection{Determination of Safety-Relevant Traffic Participants}
\label{subsecIDParticipants} The estimation of the type of road
infrastructure is followed by the determination of the
safety-relevant traffic participants in the corresponding
scenario. This forms the second level of the hierarchy. Safety
relevant traffic participants correspond to those participants
in a traffic scenario that can come close to the EGO vehicle in
the future. Hence, it is useful to predict the future behavior of
only such participants rather than all the participants in the
environment.

\begin{figure}[htb!]
    \begin{scriptsize}
    \vspace{-0.3cm}
    \centering
    \def\svgwidth{0.28\textwidth}
    \centerline{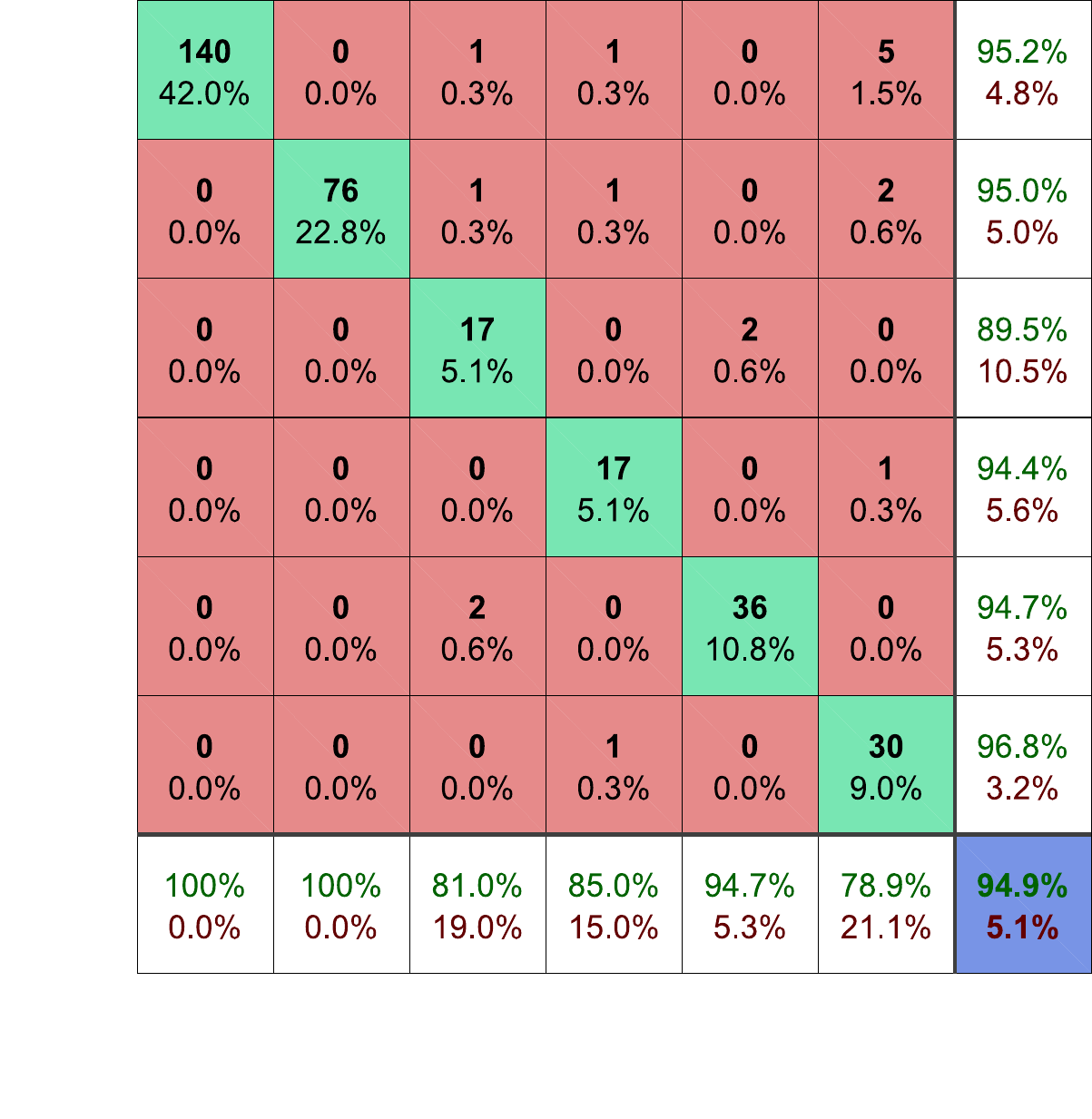}
    \end{scriptsize}
    \vspace{-0.3cm}
\def\figurename{Figure}
\caption{Confusion Matrix of the classifier to identify the constellation of the traffic participants.}
\label{figConfusionMatrix}
 \vspace{-0.4cm}
\end{figure}

In order to determine the relevant traffic participants, it is
important to determine the constellation of the participants such
as longitudinal, oncoming, crossing from left, crossing from
right, on the left, and on the right with respect to the EGO
vehicle. This can be determined based on a simple set of rules.
The classification takes into account the dynamic information
about the traffic participants and the type of road
infrastructure. Exteroceptive sensors such as radar, camera,
laserscanner, etc. are assumed to provide the information about
the traffic participants. Each traffic participant $V_\ell$ is
associated with a state vector given by
\begin{equation}
{\text{\bf{x}}}_{V_\ell} = [X_\ell,Y_\ell,v_\ell,\psi_\ell,m_\text{ego}]^\text{T},
\end{equation}
where $X_\ell$ and $Y_\ell$ correspond to position of the center
of gravity in the coordinate frame, $v_\ell$ is the absolute value
of the velocity, $\psi_\ell$ is the orientation and $m_\text{ego}$
is the slope of the EGO-lane. After the determination of the
constellations of all the traffic participants, it is necessary to
determine those which are safety-relevant. This assignment takes
into consideration the intended path of the EGO vehicle. For the
scenario shown in the Figure \ref{figScenarioAssumption}, if the
EGO vehicle (red) intends to turn right, only the traffic
participant coming from the right (blue) will be significant. If
the EGO vehicle travels straight, then both the traffic
participants (blue and green) are relevant. The free and open
traffic simulation suite SUMO~\cite{Krajzewicz12}, which
facilitates modeling traffic systems including road vehicles and
pedestrians within a realistic city infrastructure, is used for
validating the methodology.
\begin{figure}[htb!]
    \begin{scriptsize}
    \vspace{-0.2cm}
    \centering
    \def\svgwidth{0.45\textwidth}
    \centerline{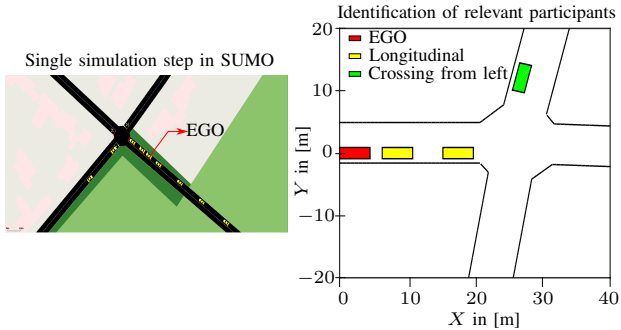}
    \end{scriptsize}
    \vspace{-0.2cm}
\def\figurename{Figure}
\caption{Validation of Hierarchical Situation Classifier using SUMO.}
\label{figvalidationSUMO}
\vspace{-0.2cm}
\end{figure}
A total of $294$ traffic scenarios with different types of road geometries are generated using the simulation environment. The scenarios were manually labeled into $K=9$ classes and $94$ scenarios are chosen in random to be the test set. The IDM with the k-NN classifier achieved a classification accuracy of $93.4\,\%$ in determining the type of road geometry. Similarly, a total of $333$ test scenarios are generated for validating the rule-based classifier to identify the $6$ different constellations of the traffic participants as described earlier. The Figure~\ref{figConfusionMatrix} shows the confusion matrix of the classifier and it has an overall accuracy of $94.9\,\%$. The results prove that the methodology is capable of reaching the correct node, i.\,e., the type of road geometry and also in identifying the
safety-relevant traffic participants. An example of a simulation step in SUMO can be seen in the left side of Figure~\ref{figvalidationSUMO}. The
information is then sent to Matlab for further processing. The
hierarchical situation classifier is able to classify the road
geometry as a $4$ way intersection and the constellations of the
traffic participants are also determined as can be seen in the
right side of Figure~\ref{figvalidationSUMO}.
\section{Predicted Occupancy Grids}
\label{secPOG}

After performing the situation classification, the next step is to
train each node of the classifier individually with the
RF algorithm to predict the future behavior
of the safety-relevant traffic participants. The future traffic
scenario includes a detailed modeling of the uncertainties
regarding the behavior of the traffic participants by considering
their multiple motion hypotheses. The probabilistic space-time
future representation of the traffic environment is the
POG and is introduced in~\cite{Nadarajan16}. For each prediction time instance
$t_\text{pred}$, a POG $\mathcal{G}_{t_{\text{pred}}}$ is
computed. Hence, over a given prediction horizon which is divided
into $\kappa$ intervals, there will be $\kappa$ POGs.
In~\cite{Nadarajan16}, an analysis on a model-based approach and
the machine-learning approach was performed for the computation of
POGs. The latter has huge advantages in terms of low computational
complexity and real-time constraints. This work introduces a
significant improvement to the existing approach by using
autoencoders to find a low-dimensional representation of the
current state of a traffic situation represented by the AOG.

\begin{figure*}[htb!]
    \begin{scriptsize}
    \vspace{0cm}
    \centering
    \def\svgwidth{0.89\textwidth}
    \hspace{1.5cm}\centerline{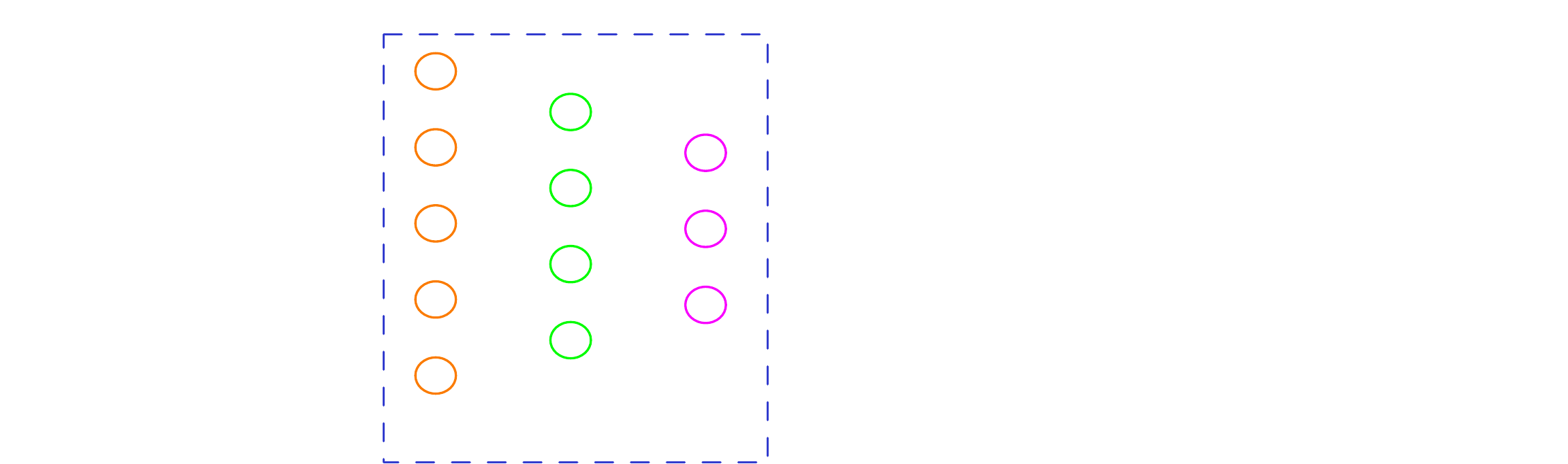}
    \end{scriptsize}
    \vspace{-0.4cm}
\def\figurename{Figure}
\caption{Estimation of Predicted Occupancy Grid.}
\label{figEstimatePOG}
 \vspace{-0.5cm}
\end{figure*}

Section~\ref{subsecAOG} describes AOGs, which are suitable
representations of the current state of traffic scenarios. AOGs
are used as inputs to the autoencoders. Section~\ref{subsecAE}
deals with the use of autoencoders for reducing the input
dimensional space and Section~\ref{subsecPOG} details the
estimation of the POGs using the reduced input dimensional space
in the RF algorithms. The outline of the machine-learning approach
adopted for the estimation of POGs can be seen in the
Figure~\ref{figEstimatePOG}.
\subsection{Augmented Occupancy Grids}
\label{subsecAOG} The future behavior of the traffic participants
depends on the intention of the drivers and the interaction
between them. Hence, information about the road infrastructure and
the dynamic information about the traffic participants is
necessary to predict the evolution of a particular scenario.
In~\cite{Nadarajan16}, an AOG $\mathcal{OG}_0$ was introduced as a
novel method to represent the current state of a traffic scenario.
The traffic scenario under observation is divided into cells of
length $\ell_\text{cell}$ and width $w_\text{cell}$ leading to $I$
columns and $J$ rows. It should be noted that for a specific
traffic scenario, there is one AOG $\mathcal{OG}_0$, where the
subscript $0$ denotes the current state at time instance $t_0$,
and there are $\kappa$ POGs $\mathcal{G}_{t_\text{pred}}$ for the
$\kappa$ prediction time instances. The cells of the occupancy
grid are augmented with additional information about the traffic
participants and the road infrastructure. The augmented attributes
correspond to the velocity, orientation, longitudinal and lateral
acceleration of a vehicle in a particular cell of the occupancy
grid. If a traffic participant $V_\ell$ with velocity $v_\ell$,
orientation $\psi_\ell$, longitudinal acceleration $a_{x,\ell}$,
lateral acceleration $a_{y,\ell}$ occupies the cell of an
occupancy grid, the attributes of the cell in $\mathcal{OG}_0$ are
$[1, v_\ell, \psi_\ell, a_{x,\ell}, a_{y,\ell}]^\text{T}$.
Similarly, the road infrastructure information is also
incorporated with the attributes of the corresponding cell being
$[1, 0, 0, 0, 0]^\text{T}$. A cell with $[0, 0, 0, 0, 0]^\text{T}$
signifies that it is unoccupied.

\subsection{Extraction of Features Using Autoencoders}
\label{subsecAE} As a result of the augmentation, the AOG
$\mathcal{OG}_{0}$ can be represented as a high dimensional
vector. For example, if an occupancy grid of dimension $80 \times
80$ is considered, the size of the $\mathcal{OG}_0$ will be $5
\times 80 \times 80$ which is equal to $32000$. The challenge is
to deal with the ``curse of dimensionality'' when
performing machine-learning tasks with high dimensional input
vectors. Hence, it is useful to extract low-dimensional meaningful
features from the high-dimensional input space in order to remove
irrelevant data, increase learning accuracy and perform better
predictions. In this work, an unsupervised technique, the stacked
denoising autoencoder is used for reducing the dimension of the
input space.

\subsubsection{Stacked Denoising Autoencoder}
\begin{figure}[htb!]
    \begin{scriptsize}
    \vspace{-0.35cm}
    \centering
    \def\svgwidth{0.4\textwidth}
    \centerline{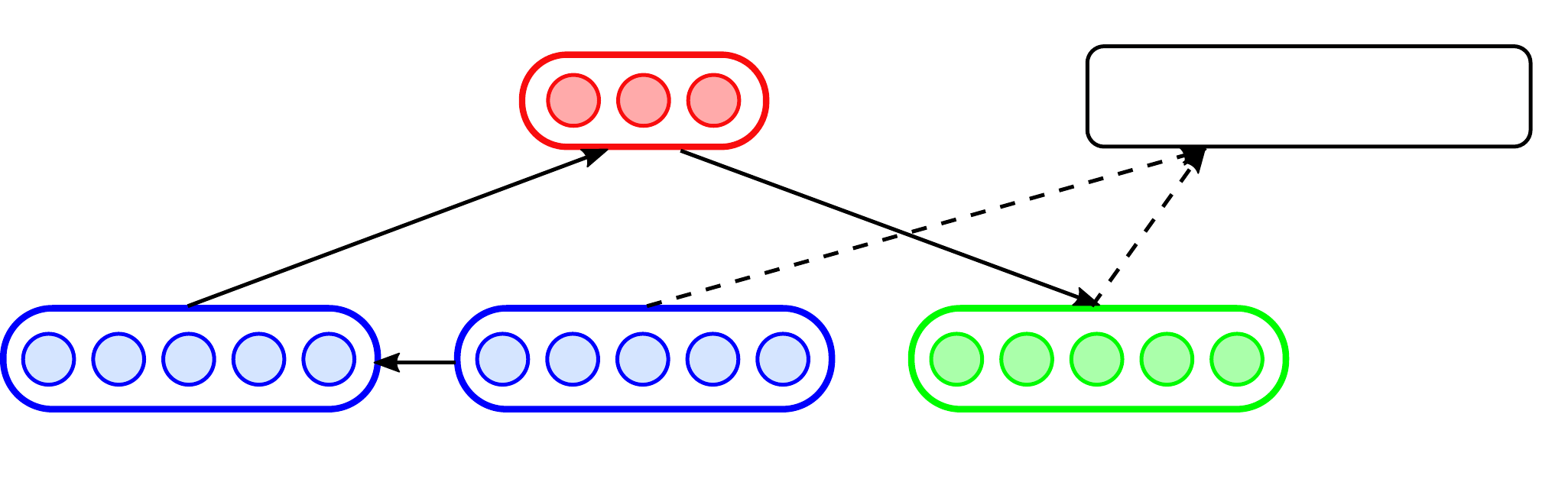}
    \end{scriptsize}
    \vspace{-0.25cm}
\def\figurename{Figure}
\caption{Denoising Autoencoder.}
\label{figStackedAutoencoders}
 \vspace{-0.3cm}
\end{figure}

An autoencoder can be considered as a neural network that is
trained to learn its input~\cite{Meng16}. It consists of three layers viz., the
input layer, hidden layer and reconstruction layer. An encoding
function maps the input data to the hidden layer and the decoding
function is responsible for mapping the hidden layer to the
reconstructed input. When the difference between the input and the
reconstructed input is minimal, the hidden layer vector can be
stated as a low-dimensional representation of the input. In order
to prevent the autoencoders from learning the identity function
and to improve their ability to capture important representations,
a denosing autoencoder is used. In~\cite{Vincent08}, it was shown
that better representations can be learnt when using the SDA. 
The SDA consists of multiple denoising autoencoders stacked one 
above the other, where the output of each layer is fed in as 
input to the successive layer. A greedy layer-wise training 
procedure is adopted in the case of SDA.
Figure~\ref{figStackedAutoencoders} shows a single layer of the 
SDA model.

Let $l=1, \ldots, nl$ correspond to the layer number of the SDA.
The $l$-th layer visible vector, hidden vector and reconstructed
vector are represented as $\boldsymbol{p}_g^{(l)}$,
$\boldsymbol{q}_g^{(l)}$ and $\boldsymbol{r}_g^{(l)}$
respectively, where $g=1,\ldots, G$ with $G$ denoting the total number of
training data. The $g$-th AOG $\mathcal{OG}_{0,g}$ is represented
as the vector $\boldsymbol{p}_g^{(1)}\in \mathbb{R}^{5 \cdot I
\cdot J}$. The denoising autoencoder is constructed by adding
noise to $\boldsymbol{p}_g^{(l)}$ to create a partially destroyed
version of the input $\boldsymbol{\tilde{p}}_g^{(l)}$ by
stochastic mapping
~\cite{Vincent08}. The three types of commonly used corrupting
operations are Gaussian noise, masking noise, and salt and pepper
noise~\cite{Meng16}. In this work, Gaussian noise is used for the
corrupting operation. After performing the corruption operation,
the $l$-th hidden layer vector $\boldsymbol{q}_g^{(l)}$ is
constructed using the encoding function
$h_\theta^{(l)}(\boldsymbol{\tilde{p}}_g^{(l)})$:
\begin{equation}
\boldsymbol{q}_g^{(l)} = h_\theta^{(l)}(\boldsymbol{\tilde{p}}_g^{(l)}) = f\Bigl(\boldsymbol{W}^{(l)}\boldsymbol{\tilde{p}}_g^{(l)} + \boldsymbol{b}^{(l)}\Bigr),
\end{equation}
where $\theta^{(l)} =
\{\boldsymbol{W}^{(l)},\boldsymbol{b}^{(l)}\}$, with
$\boldsymbol{W}^{(l)}$ and $\boldsymbol{b}^{(l)}$ being the weight
matrix and bias vector of the $l$-th layer, respectively. The
function $f(\cdot)$ corresponds to the activation function such as
sigmoid, linear, hyperbolic tangent, etc. From the hidden layer,
the decoding function $s_{\theta'}^{(l)}(\boldsymbol{q}_g^{(l)})$
is used to obtain the reconstructed input vector
$\boldsymbol{r}_g^{(l)}$ as follows:
\begin{equation}
\boldsymbol{r}_g^{(l)} = s_{\theta'}^{(l)}(\boldsymbol{q}_g^{(l)}) = f\Bigl(\boldsymbol{W}^{(l)'}\boldsymbol{q}_g^{(l)}+\boldsymbol{b}^{(l)'}\Bigr),
\end{equation}
where $\theta^{(l)'} =
\{\boldsymbol{W}^{(l)'},\boldsymbol{b}^{(l)'}\}$. In this paper,
tied weights and bias are used, i.\,e.,
$\boldsymbol{W}=\boldsymbol{W}'$ and
$\boldsymbol{b}=\boldsymbol{b}'$, respectively. The $l$-th layer
loss function $\mathcal{L}^{(l)}$ for the reconstruction of the
input is the second-order loss function with a regularization
parameter to avoid overfitting and is given by
\begin{align}
\mathcal{L}^{(l)}(\boldsymbol{W}^{(l)},\boldsymbol{b}^{(l)})=\frac{1}{2G}\sum_{g=1}^{G}\|\boldsymbol{r}_g^{(l)}-\boldsymbol{p}_g^{(l)}\|^2\\
\qquad \qquad \; +\frac{\lambda \nonumber
}{2}\sum_{l=1}^{nl}\sum_{x=1}^{sl}\sum_{y=1}^{sl+1}\Bigl(\boldsymbol{W}_{xy}^{(l)}\Bigr),
\end{align}
where $\lambda$ is the weight decay parameter, $sl$ represents the
number of units on the $l$th layer and $\boldsymbol{r}_g^{(l)}$ is
a function of the weights $\boldsymbol{W}^{(l)}$ and the bias
$\boldsymbol{b}^{(l)}$. Thus, the final output of the SDA for the
$g$-th vector is $\boldsymbol{q}_g^{(nl)}$.


\subsection{Estimation of POGs Using Random Forest}
\label{subsecPOG}
\begin{figure}[htb!]
    \begin{scriptsize}
    \vspace{-0.2cm}
    \hspace{-0.1cm}
    \centering
    \def\svgwidth{0.23\textwidth}
    \centerline{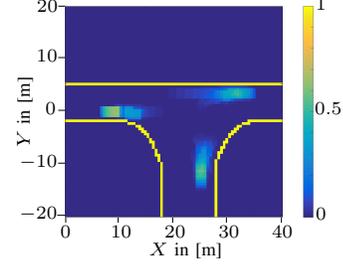}
    \end{scriptsize}
    \vspace{-0.5cm}
\def\figurename{Figure}
\caption{Predicted Occupancy Grid for $t_\text{pred}=2.0$s with the color bar denoting the probability of occupancy $\text{p}_{t_\text{pred}}^{ij}$.}
\label{figPOG}
\vspace*{-0.5cm}
\end{figure}

With the reduction of the high-dimensional input space $\mathcal{OG}_0$ to low-dimensional meaningful features $\boldsymbol{q}^{(nl)}$,  it is now required to estimate the POGs $\mathcal{G}_{t_\text{pred}}$. The RF algorithms are responsible for performing the mappings,
\begin{equation}
\boldsymbol{q}^{(nl)}\mapsto\mathcal{G}_{t_\text{pred}}.
\end{equation}
The main reasons for using the RF algorithm in this paper are its well known properties such as: good generalization, low number of hyper-parameters to be tuned during training and good performance with high-dimensional data. Also, faster predictions of output is feasible due to parallel processing. The POG is of the dimension $I \times J$ and let $g_{t_\text{pred}}^{ij}$ denote the $(i,j)$-th cell of the POG at prediction instance $t_\text{pred}$. The probability of occupancy of $g_{t_\text{pred}}^{ij}$ at $t_\text{pred}$ is $\text{p}_{t_\text{pred}}^{ij}$. The probability $\text{p}_{t_\text{pred}}^{ij}$ depends on the probabilities of the multiple trajectories of the traffic participants in a traffic scenario. It is also important to note that multiple traffic participants trajectory hypotheses can simultaneously occupy a cell of the POG. However, the maximum probability of $g_{t_\text{pred}}^{ij}$ is limited to 1. Thus,
\begin{equation}
\text{p}_{t_\text{pred}}^{ij}=
\text{min}\left(1,\sum_{\ell=1}^{L}\left(\left(\boldsymbol{z}_{\text{V}_\ell,t_\text{pred}}^{ij}\right)^T\textbf{p}(h_{{\text{V}_\ell},t_\text{pred}})\right)\right),
\end{equation}
where $\boldsymbol{z}_{\text{V}_\ell,t_\text{pred}}^{ij}$ corresponds to a binary vector of size $S$, where S is the number of trajectory hypotheses per traffic participant $V_\ell$. It takes up values $0$ or $1$ depending on the occupancy of the $S$ multiple hypotheses trajectories of the traffic participant $V_\ell$. $\textbf{p}(h_{{\text{V}_\ell},t_\text{pred}})$, also a vector of size $S$, comprises of the probabilities of the $S$ multiple hypotheses at prediction instance $t_\text{pred}$. Since $\text{p}_{t_\text{pred}}^{ij}$ is a continuous value between 0 and 1, the regression task using the RF is performed. Also, the cells of the POG are assumed to be independent of each other and to predict the probability of each cell $\text{p}_{t_\text{pred}}^{ij}$, one RF is trained per $g_{t_\text{pred}}^{ij}$. Thus, a set of trained RFs $\bigl\{\text{RF}_{t_{\text{pred}}}^{11}, \ldots, \text{RF}_{t_{\text{pred}}}^{IJ} \bigr\}$ exist for a particular $t_\text{pred}$ to estimate the POG $\mathcal{G}_{t_\text{pred}}$. A pictorial representation of the methodology can be seen in the Figure~\ref{figEstimatePOG}. An example of the POG for $t_\text{pred} = 2.0$s with three traffic participants can be seen in the Figure~\ref{figPOG}. The color bar denotes the probability of occupancy $\text{p}_{t_\text{pred}}^{ij}$. The cells of the POG occupied by the road infrastructure have an occupancy value of 1.

\section{Simulations and Experiments}
\label{secExp}
Simulations are performed in order to validate two aspects
of the proposed methodology, namely, the ability
of the SDA to achieve dimensionality reduction on the AOG and the quality of the predicted POGs using the low-dimensional features extracted from the AOG. Results from the experiments carried out with real vehicles at an outdoor test facility are also shown. 

\subsection{Generation of Data}
\begin{figure}[htb!]
    \begin{scriptsize}
    \vspace{-0.2cm}
    \centering
    \def\svgwidth{0.21\textwidth}
    \centerline{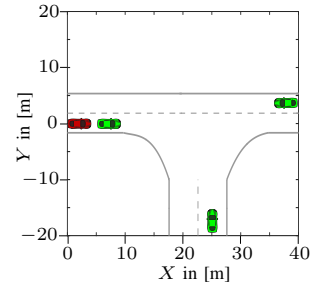}
    \end{scriptsize}
    \vspace{-0.2cm}
\def\figurename{Figure}
\caption{Scenario under consideration.}
\label{figScenario}
\vspace{-0.4cm}
\end{figure}

With the aim of validating the methodology using the SDA and RFs, only a particular class of the hierarchical situation classifier is considered to perform the simulations. Hence, a three way intersection with multiple traffic participants on a span of ${40\text{m} \times 40}$m is chosen as the traffic scenario as can be seen in the Figure~\ref{figScenario}. The red vehicle corresponds to the EGO vehicle and the green vehicles are the traffic participants in the environment. It is important to note that the behavior of only the traffic participants is predicted and not the EGO vehicle. This is because the behavior of the traffic participants cannot be influenced whereas the behavior of the EGO vehicle can. 

The grid resolution of the POG is chosen to be $0.5$m thereby resulting in $I=80$ and $J=80$. The maximum longitudinal acceleration and deceleration of the traffic participants considered during the generation of multiple hypotheses are $4.5$m/s$^2$ and $9.0$m/s$^2$, respectively. The maximum lateral acceleration is $7.0$m/s$^2$.

A total of 2850 initial states, i.\,e. $\mathcal{OG}_0$, of the above mentioned traffic scenario is generated by varying the number of traffic participants in the environment, their respective positions, velocities and longitudinal accelerations. The initial states $\mathcal{OG}_0$ and the ground truth output $\mathcal{G}_{t_\text{pred}}$ are generated with the help of a model-based approach as mentioned in~\cite{Nadarajan16}.
The data is generated with a combination of $3, 2$ and only $1$ traffic participant in the environment. The velocity is varied between $10$km$/$h and $50$km$/$h. The position of the traffic participants is changed over a range of $10$m. The variations in the longitudinal acceleration are about $2.5$m$/$s$^2$. The prediction time instance $t_\text{pred}$ is chosen to be $2.0$s. Since the machine-learning models have to be validated, a total of $1950$ traffic scenarios which is approximately two-thirds of the total traffic scenarios is chosen for the training and the remaining $900$ scenarios are chosen to be the test set.
\subsection{Quality Metrics}
In order to validate the methodology and to determine the quality of the trained machine-learning models, it is important to introduce appropriate quality metrics. In this work, two quality metrics are used. The first is to quantify the capability of the low-dimensional features learned using the SDA to reconstruct the high-dimensional input. The second quality metric is to measure the ability of the RFs to predict the POGs.
\subsubsection{Quality of Feature Extraction}
The quality of a trained SDA depends on its ability to reconstruct the given input vector. The deviation between the original input vector and its corresponding reconstructed vector can be used as a quantity to ascertain the quality of dimensionality reduction. Hence, the \emph{Root Mean Squared Error} (RMSE) is used as the metric to compare the similarity. Using the notations introduced in the Section~\ref{subsecAE}, the error for one AOG is given by, 
\begin{align}
\label{eqMSE1}
\varepsilon = \|\boldsymbol{r}^{(1)}-\boldsymbol{p}^{(1)}\|.
\end{align}
\subsubsection{Quality of POG Prediction}
In~\cite{Nadarajan16}, a quality metric was defined to quantify the prediction accuracy of the POGs using the RF algorithm. The introduced measure was strict with respect to not rewarding the estimation of free spaces in the POGs.  The ground truth and the estimated POG for the prediction time instance $t_\text{pred}$ is given by $\mathcal{G}_{t_\text{pred}}$ and $\hat{\mathcal{G}}_{t_\text{pred}}$. Since the quality measure does not account for the estimation of free spaces, only the non-empty cells of the POG are considered. Let $\mathcal{B}$ and $\mathcal{D}$ denote the set of cells with non-zero values in the $\mathcal{G}_{t_\text{pred}}$ and $\hat{\mathcal{G}}_{t_\text{pred}}$, respectively. The cardinality $\mathcal{K}$ of the set $\left(\mathcal{B}\cup\mathcal{D}\right)\setminus
\left(\mathcal{B}\cap\mathcal{D}\right)$ is given by,
\begin{align}
\mathcal{K}=\left\vert \left(\mathcal{B}\cup\mathcal{D}\right)\setminus
\left(\mathcal{B}\cap\mathcal{D}\right) \right\vert.
\end{align}
Thus, the quality measure $\epsilon_{t_\text{pred}}$ for the prediction time instance $t_\text{pred}$ for one POG is defined as
\begin{align}
\label{eqMSE2}
\epsilon_{{t_\text{pred}}}=\sqrt{\frac{1}{\mathcal{K}}\sum\limits_{i=1}^{I}\sum\limits_{j=1}^{J}\Big(\hat{\text{p}}_{t_\text{pred}}^{ij}-\text{p}_{t_\text{pred}}^{ij}\Big)^2},
\end{align}
where $\hat{\text{p}}_{t_\text{pred}}^{ij}$ and $\text{p}_{t_\text{pred}}^{ij}$ are the probabilities stored in the $(i,j)$-th cell of the estimated and ground truth POG, respectively.
\subsection{Simulation Results}
The results of the simulation are presented in this section.
\subsubsection{Results of Dimensionality Reduction}
This part presents the results with respect to the dimensionality reduction performed using the SDA. The SDA from the Matlab Toolbox for Deep Learning~\cite{Palm12} is used in this work. The number of layers $nl$ in the SDA is chosen to be $3$. The number of hidden units in the first, second and third layer of the SDA are $2000$, $1000$ and $500$, respectively. Thus, the input space $\mathcal{OG}_0$ with a dimension of $32000$ is reduced to a low-dimensional feature vector $\boldsymbol{q}^{(nl)}$ of size $500$. The corrupting operation employed in this work is Gaussian Noise with the noise level of the SDA being $0.3$. The learning rate of all the layers of the SDA is chosen to be $0.001$. The weight decay parameter $\lambda$ and the momentum are assigned $0.005$ and $0.9$, respectively. The maximum number of iterations is restricted to $400$. The above mentioned hyperparameters of the SDA are chosen according to~\cite{Meng16} and~\cite{Vincent08}, where a detailed analysis on the effect of each hyperparameter on the performance of dimensionality reduction was done.
\begin{figure}[htb!]
    \begin{scriptsize}
    \vspace{-0.2cm}
    \centering
    \def\svgwidth{0.15\textwidth}
    \centerline{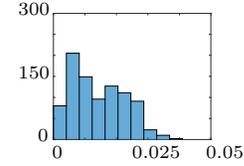}
    \end{scriptsize}
    \vspace{-0.2cm}
\def\figurename{Figure}
\caption{Histogram of $\varepsilon$ for 900 test scenarios.}
\label{fighistMSE}
\vspace{-0.25cm}
\end{figure}
The Figure~\ref{fighistMSE} shows the histogram of the error $\varepsilon$ computed according to the Equation~(\ref{eqMSE1}). The average RMSE $\bar{\varepsilon}$ over the 900 test samples is $0.0143$. The range of the values in the AOG is between $-7$ and $13$ as it includes the information starting from the occupancy to the dynamic information of the traffic participants. The average absolute value per cell of the AOG over all the test scenarios is $0.112$. 
The results prove that the low-dimensional feature $\boldsymbol{q}^{(nl)}$ extracted using the $3$-layered SDA is a robust representation of the high-dimensional input.

\subsubsection{Results of POG Prediction}
\begin{figure}[htb!]
    \begin{scriptsize}
    \vspace{0cm}
    \centering
    \def\svgwidth{0.48\textwidth}
    \centerline{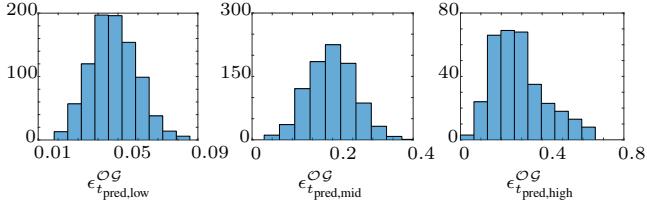}
    \end{scriptsize}
    \vspace{0cm}
\def\figurename{Figure}
\caption{Histogram of $\epsilon_{t_\text{pred}}^{\mathcal{OG}}$ for 900 test scenarios.}
\label{fighistTwoSecOrg}
\vspace{-0.4cm}
\end{figure}

\begin{figure}[htb!]
    \begin{scriptsize}
    \vspace{0cm}
    \centering
    \def\svgwidth{0.48\textwidth}
    \centerline{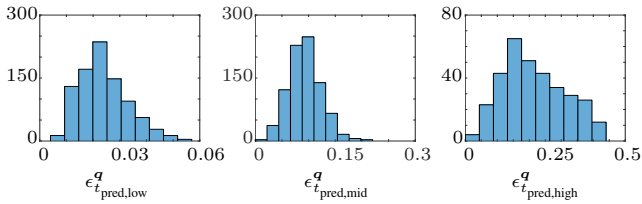}
    \end{scriptsize}
    \vspace{0cm}
\def\figurename{Figure}
\caption{Histogram of $\epsilon_{t_\text{pred}}^{\boldsymbol{q}}$ for 900 test scenarios.}
\label{fighistTwoSecRed}
\vspace{-0.4cm}
\end{figure}

\begin{figure}[htb!]
    \begin{scriptsize}
    \vspace{0cm}
    \centering
    \def\svgwidth{0.48\textwidth}
    \centerline{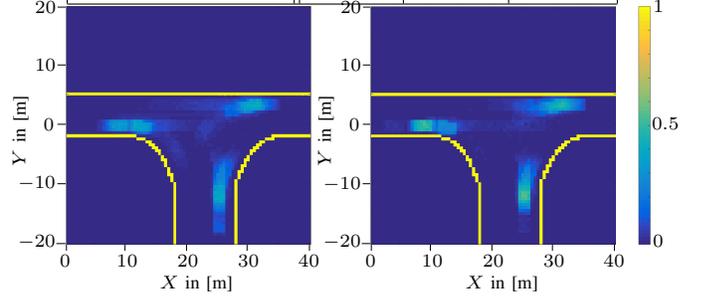}
    \end{scriptsize}
    \vspace{-0.2cm}
\def\figurename{Figure}
\caption{Predicted Occupancy Grids using $\mathcal{OG}_0$ and $\boldsymbol{q}^{(nl)}$ as input to RFs  with the color bar denoting the probability of occupancy $\text{p}_{t_\text{pred}}^{ij}$.}
\label{figPOGPred}
\vspace{-0.5cm}
\end{figure}

The simulation results for the estimation of POGs $\mathcal{G}_{t_\text{pred}}$ using the RFs with prediction time instance $t_\text{pred}=2.0$s is presented in this section. It is important to realize whether the process of dimensionality reduction has increased the learning accuracy. Hence, two sets of RFs are trained for the prediction of the POG, one using the original high-dimensional $\mathcal{OG}_0$ as the input and the other using the extracted low-dimensional feature $\boldsymbol{q}^{(nl)}$ as the input. The error $\epsilon_{t_\text{pred}}$ is computed separately for both the RF models using the Equation~(\ref{eqMSE2}). Let $\epsilon_{t_\text{pred}}^{\mathcal{OG}}$ and $\epsilon_{t_\text{pred}}^{\boldsymbol{q}}$ be the error computed for the RFs trained using $\mathcal{OG}_0$ and $\boldsymbol{q}^{(nl)}$ as their input respectively and their corresponding histograms for the $900$ test scenarios can be seen in the Figure~\ref{fighistTwoSecOrg} and~\ref{fighistTwoSecRed}. For better interpretation of the results, three error estimates $\epsilon_{t_\text{pred},\text{low}}$, $\epsilon_{t_\text{pred},\text{mid}}$ and $\epsilon_{t_\text{pred},\text{high}}$ are computed for low, mid and high values of the probability $\text{p}_{t_\text{pred}}^{ij}$, respectively. The range of the probability $\text{p}_{t_\text{pred}}^{ij}$ for the computation of $\epsilon_{t_\text{pred},\text{low}}$ is $[0,0.25]$. Similarly, for the computation of $\epsilon_{t_\text{pred},\text{mid}}$ and $\epsilon_{t_\text{pred},\text{high}}$, the range of the probability are $(0.25,0.75]$ and $(0.75,1.0]$, respectively.
The average error $\bar{\epsilon}$ estimated over the $900$ test scenarios for both the RF models can be seen in the Table~\ref{tabelResults}. The first row of the Table~\ref{tabelResults} contains the mean error estimates computed for the RFs trained using the original input dimension $\mathcal{OG}_0$ and the second row contains the mean error estimates of the RFs trained using $\boldsymbol{q}^{(nl)}$. By comparing the results of the two RF models, it can be clearly seen that the mean error is reduced by approx. $50\,\%$ for both the low and mid occupancy values. Even though an occurrence of high probability is unlikely when considering a prediction horizon of $2.0$s, the error is reduced by approx. $25\,\%$. It should also be noted that the dimensionality reduction minimizes the under or over estimation of probabilities. This validates that performing efficient dimensionality reduction on high-dimensional input space helps in elimination of noise, increasing learning accuracy and thereby performing better predictions. The time required for the training of the RFs is also significantly reduced with the RFs considering lesser dimensions for finding the best split during the learning process. An example of the POG $\mathcal{G}_{t_\text{pred}}$, for the traffic scenario shown in the Figure~\ref{figScenario}, estimated using the $\mathcal{OG}_0$ and $\boldsymbol{q}^{(nl)}$ is shown in the left and right side of the Figure~\ref{figPOGPred}, respectively and their corresponding errors $\epsilon_{t_\text{pred}}$ are $0.0415$ and $0.0249$, respectively. The ground truth for the estimated POG can be seen in the Figure~\ref{figPOG}. It is also important to note that the proposed methodology for the estimation of POGs is capable of predicting the behavior of the traffic participants even if the number of the traffic participants are varying. In the simulations performed, the traffic scenarios had $3$, $2$ and only $1$ traffic participant and the machine-learning approach is able to capture this information and perform the predictions accordingly. 

\begin{table}[!t]
\renewcommand{\arraystretch}{1.3}
\caption{Comparison of the errors using the original and reduced input dimension for $t_\text{pred}=2.0\text{s}$}
\label{tabelResults}
\centering
\begin{tabular}{|c||c|c|c|}
\hline
Input to the RF model & $\bar{\epsilon}_{t_\text{pred},\text{low}}$ & $\bar{\epsilon}_{t_\text{pred},\text{mid}}$ & $\bar{\epsilon}_{t_\text{pred},\text{high}}$  \\
\hline
$\mathcal{OG}_0$ & $0.0478$ & $0.1967$ & $0.2855$\\
\hline
$\boldsymbol{q}^{(nl)}$ & $0.0239$ &  $0.0930$ & $0.2156$ \\
\hline
\end{tabular}
\vspace{-0.5cm}
\end{table}

\vspace{-0.2cm}
\subsection{Experiments with Real Vehicles}
\begin{figure}[htb!]
    \begin{scriptsize}
    \vspace{-0.2cm}
    \centering
    \def\svgwidth{0.35\textwidth}
    \centerline{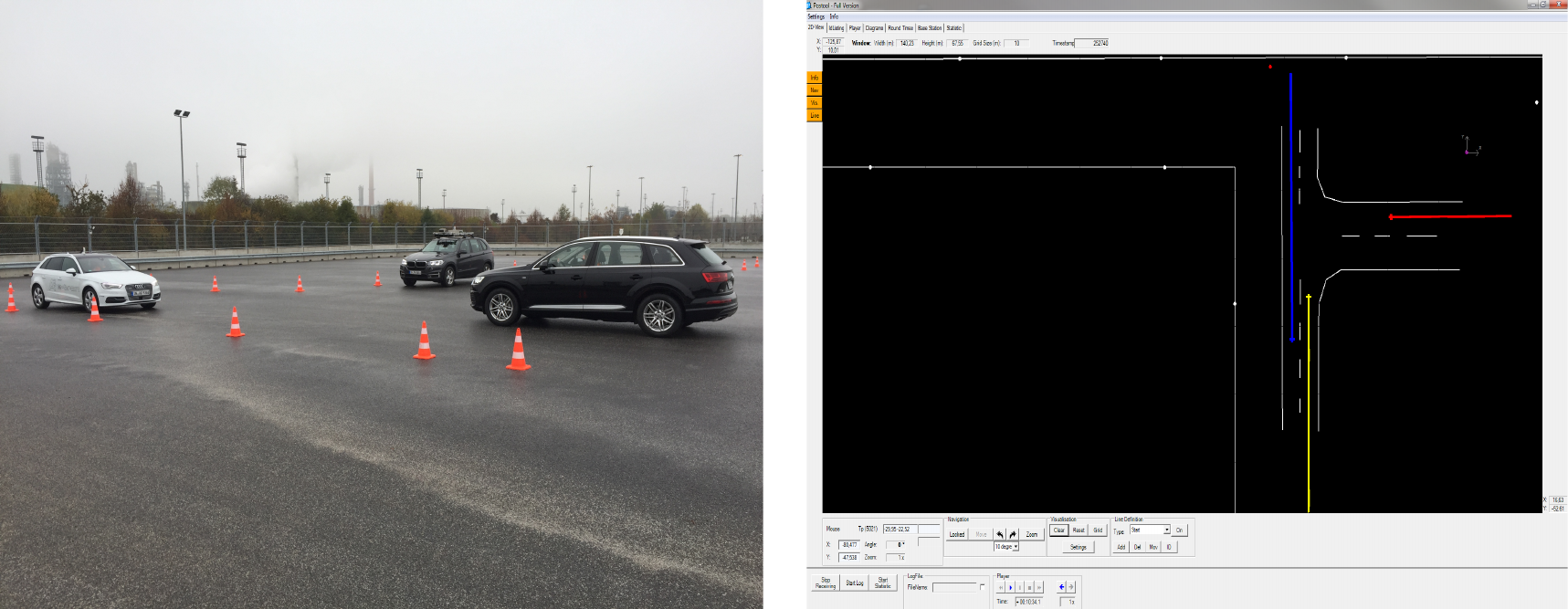}
    \end{scriptsize}
    \vspace{-0.1cm}
\def\figurename{Figure}
\caption{Experiments with real vehicles at the outdoor test facility.}
\label{figExp}
\vspace*{-0.2cm}
\end{figure}

Experiments are carried out with real vehicles at the Center of Automotive Research on Integrated Safety Systems and Measurement Area (CARISSMA) outdoor test facility of Technische Hochschule Ingolstadt to determine the plausibility of the approach. The test track with the experimental vehicles can be seen in the left side of the Figure~\ref{figExp}.

\begin{figure}[htb!]
    \begin{scriptsize}
    \vspace{-0.2cm}
    \centering
    \def\svgwidth{0.47\textwidth}
    \centerline{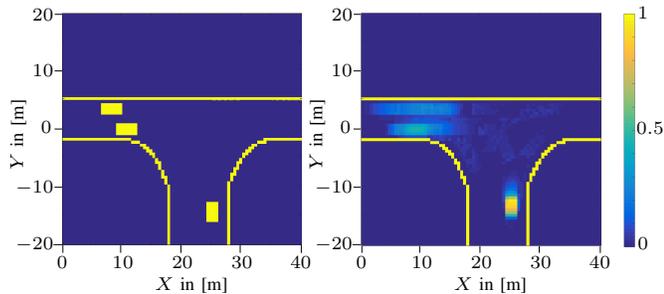}
    \end{scriptsize}
    \vspace{-0.2cm}
\def\figurename{Figure}
\caption{Reference occupancy grid and the estimated Predicted-Occupancy Grid $\hat{\mathcal{G}}_{t_\text{pred}}$ with the color bar denoting the probability of occupancy $\text{p}_{t_\text{pred}}^{ij}$.}
\label{figResultExp}
\vspace*{-0.4cm}
\end{figure}

A set of non-critical test scenarios from the simulation environment is selected at random and the maneuvers of the corresponding scenarios are performed at the outdoor test facility to evaluate the performance of the machine-learning approach in predicting the behavior of the traffic participants. The reference state information of the traffic participants is provided by a \emph{Local Position Measurement} (LPM) System~\cite{Stelzer04}. The real-time tracking of the vehicles can be visualized with the help of the PosTool software and one such visualization can be seen in the right side of the Figure~\ref{figExp}. The blue, red and yellow lines correspond to the trajectories of the traffic participants. The information from the LPM system is imported into the Matlab environment to perform further analysis. For the scenario performed at the test track, the occupancy grid for prediction time instance $t_\text{pred}=2.0$s is estimated. Additionally, a reference occupancy grid is available by using the LPM measurements at time $t_0+2.0$s, where $t_0$ corresponds to the start of the scenario. This reference occupancy grid is then compared with the estimated POG $\hat{\mathcal{G}}_{t_\text{pred}}$ computed using the machine-learning approach. The reference occupancy grid computed using the measurements from the LPM measurement and the estimated POG $\hat{\mathcal{G}}_{t_\text{pred}}$ can be seen in the left and right side of the Figure~\ref{figResultExp}, respectively. It is important to note that the training process of the RFs is based only on the simulation data. Also with respect to the reference occupancy grid, there is no uncertainty regarding the behavior of the traffic participants. This is because it does not involve any prediction and is determined only by measuring the exact position of the traffic participants at $t_0+2.0$s. As can be seen in the Figure~\ref{figResultExp}, the position of the traffic participants in the reference occupancy grid matches with the region of the $\hat{\mathcal{G}}_{t_\text{pred}}$ which has high probability of occupancy. This demonstrates that the machine-learning approach is capable of predicting the behavior of the traffic participants under real-world conditions provided the required information is available from the sensors.
\section{An application in Vehicle Safety}
\label{secApplication}
The real time capability of the machine-learning approach to estimate the POGs finds application in the field of vehicle safety. The detailed modeling of the uncertainties regarding the motion behavior of the other traffic participants helps in improving components of vehicle safety, such as criticality estimation, trajectory planning, etc.  Under critical situations, it is important to plan a trajectory for the EGO vehicle which has a very low risk of collision with the surrounding traffic participants.  Let $u = 1, \ldots, U$, with $U$ being the number of maneuverable trajectories by the EGO vehicle over the prediction time horizon $t_\text{pred}$. The $t_\text{pred}$ is divided into $\kappa$ intervals thereby resulting in $\kappa$ POGs. Each maneuver of the EGO vehicle will result in a different occupancy in the $\kappa$ POGs. Let $c_{u,t_\text{pred}}$ be the sum of the probabilities of the cells of the POG $\mathcal{G}_{t_\text{pred}}$ which are simultaneously occupied by the $u$-th trajectory of the EGO vehicle at prediction instance ${t_\text{pred}}$. Hence, the number of $c_{u,t_\text{pred}}$ computed will be $\kappa$. Thus, the trajectory with 
$\min_{u}\{\max_{t_\text{pred}}\{c_{u,t_\text{pred}}\}\}$ will be the safe trajectory for the EGO vehicle, as it has the least probability of collision with the surrounding traffic participants. Analysis of this approach is currently being carried out.

\section{Conclusion}
This paper presents a methodology for predicting the evolution of different kinds of traffic scenarios by including the uncertainties regarding the motion behavior of the traffic participants. A hierarchical situation classifier is used to classify the different traffic scenarios based on road geometry and safety-relevant traffic participants, and a set of Random Forests are individually trained for each class of the classifier to predict the traffic scenario. The Image Distortion Model and a set of predefined rules are used as the decision process in the classifier. Simulations are carried out in the SUMO-Matlab environment to validate the classifiers and the results are promising. The unsupervised dimensionality reduction using Stacked Denoising Autoencoders is performed on the Augmented Occupancy Grid. The low-dimensional features are capable of increasing the learning and prediction accuracy of the Random Forests. They also contribute towards a significant reduction in the time required for the training of the Random Forests. The results of the simulation using the $900$ test scenarios and the experiments using real vehicles prove that the proposed machine-learning approach is capable of predicting a reliable estimate of the Predicted-Occupancy Grid. An application of the Predicted-Occupancy Grids in planning safe trajectories for the EGO vehicle under safety critical situations is also presented. 

Future work will focus on the use of a convolutional autoencoder for the dimensionality reduction and in demonstrating applications of Predicted-Occupancy Grids for vehicle safety.  
\vspace*{-0.4cm}

\section*{Acknowledgment}
\vspace*{-0.1cm}
The authors would like to thank the CARISSMA team for their help in performing the experiments with real vehicles at the outdoor test facility. 
\vspace*{-0.1cm}


%

\end{document}